\documentclass[10pt,a4paper]{article}

\usepackage{natbib}
\usepackage{makeidx}  
\usepackage{url} 
\usepackage{color}
\usepackage{amsfonts}
\usepackage{amssymb}
\usepackage{amsmath}
\usepackage{xfrac}
\usepackage[misc]{ifsym}
\usepackage[absolute]{textpos}
\usepackage[colorlinks]{hyperref}
\usepackage{cleveref}

\allowdisplaybreaks[4]
\let\originalleft\left
\let\originalright\right
\renewcommand{\left}{\mathopen{}\mathclose\bgroup\originalleft}
\renewcommand{\right}{\aftergroup\egroup\originalright}

\newcommand{\B}{\mathbf{B}}

\newcommand{\uu}{\mathbf{u}}
\newcommand{\U}{\mathbf{U}}
\newcommand{\vv}{\mathbf{v}}
\newcommand{\V}{\mathbf{V}}
\newcommand{\w}{\mathbf{w}}
\newcommand{\W}{\mathbf{W}}

\newcommand{\X}{\mathbf{X}}

\newcommand{\z}{\mathbf{z}}
\newcommand{\Z}{\mathbf{Z}}

\newcommand{\N}{\mathcal{N}}

\newcommand{\tensorA}{\mathcal{A}}

\newcommand{\tensorW}{\mathcal{W}}
\newcommand{\tensorX}{\mathcal{X}}

\newcommand{\tensorXWH}{\widehat{\mathcal{X}}}

\newcommand{\eps}{\boldsymbol{\epsilon}}
\newcommand{\Ta}{\boldsymbol{\tau}}

\newcommand{ \outerProd}{\circ} 
\newcommand{\modeKProd}{\times_{2}}
\newcommand{\modeKProdOne}{\times_{1}}

\newcommand{\redcomment}[1]{\textcolor{black}{#1}} 

\begin{document} 

\begin{textblock}{14}(2,2)
\noindent\Large \redcomment{Citation Info: \\ Khan, S.A., Lepp\"aaho, E. and Kaski, S., 2016.  Bayesian multi-tensor factorization, \textit{Machine Learning}, 105(2) 233-253\\ \href{http://dx.doi.org/10.1007/s10994-016-5563-y}{doi:10.1007/s10994-016-5563-y}}
\end{textblock}

\title{Bayesian multi-tensor factorization
}


\author{Suleiman A. Khan  \and
        Eemeli Lepp\"aaho \and
        Samuel Kaski
}

%
%
%

\maketitle

\maketitle
\begin{centering}
$^{1}$ Helsinki Institute for Information Technology HIIT, Department of Computer Science, Aalto University, Finland \\
$^{2}$ Helsinki Institute for Information Technology HIIT, Department of Computer Science, University of Helsinki, Finland \\
\end{centering}

\begin{abstract} 
We introduce Bayesian multi-tensor factorization, a model that is the
first Bayesian formulation for joint factorization of multiple
matrices and tensors. The research problem generalizes the joint matrix-tensor
factorization problem to arbitrary sets of tensors of any depth,
including matrices, can be interpreted as unsupervised multi-view
learning from multiple data tensors, and can be generalized to relax the
usual trilinear tensor factorization assumptions. The result is a
factorization of the set of tensors into factors shared by any subsets
of the tensors, and factors private to individual tensors.  We
demonstrate the performance against existing baselines in multiple
tensor factorization tasks in structural toxicogenomics and functional
neuroimaging.

\end{abstract}


\section{Introduction} \label{sec:introduction}

Matrix and tensor factorization methods have been studied for
data analysis and machine learning for decades. These methods 
decompose a single data set into a low-dimensional representation of factors that explain the variation 
in it. With linked data sets becoming increasingly common, joint 
factorization of multiple data sources is now gaining significant attention.

Joint factorization of multiple matrices integrates information 
from multiple coupled data sets. 
It decomposes them into underlying latent components or factors, 
taking advantage of the common structure between all of them. 
For the simplest case of two paired matrices, 
canonical correlation analysis finds latent variables that capture the shared 
variation explaining them \citep{Bach05,Hardoon04,Hotelling36}.
While canonical correlation analysis searches for 
common patterns between two data matrices, its straightforward 
extensions have limited applicability
in multiple coupled matrices. Recently, a multi-view method called group factor 
analysis \citep[GFA;][]{Klami14gfa,Virtanen12},
has been presented for decomposing multiple paired matrices. 
GFA decomposes multiple coupled matrices identifying 
both the co-variation patterns shared between some of the data sets, as well as 
those specific to each.

Tensor factorizations have also been considered as a means of 
analyzing multiple matrices by coupling them together as slabs of 
a tensor. These factorizations are more general and are 
able to take advantage of the natural tensor structure of the data.
A host of low-dimensional tensor factorization methods have been 
proposed earlier \citep[see][for a review]{KoldaBader09}. The most well-known are the 
CANDECOMP/PARAFAC \citep[CP;][]{Carroll70,Harshman70} and the 
Tucker family of models \citep{kiers1991hierarchical,Tucker66}. 
CP assumes a trilinear structure in the data and is easier to interpret, 
while the Tucker family defines more generic models for 
complex interactions.

However, neither the tensor factorization nor the joint matrix factorization 
is able to factorize mixed and partially linked data sets. 
Recently, fusion of partially coupled data sets has been 
discussed, for example to predict the values in a tensor with side information
from a matrix, or vice versa. For example, \cite{Acar13} 
used metabolomics data of fluorescence emission $\times$ excitation 
measurements and NMR recordings of 
several human blood samples to form a coupled tensor and a matrix, 
to demonstrate that joint factorization outperforms individual factorization.
The concept of such multi-block 
decompositions was originally introduced by \cite{smilde00}, 
and proposed by \cite{harshman1994parafac}, though the recent formulation by \cite{Acar11,Acar13} has brought 
coupled matrix tensor factorizations to practical use.

We call this general research problem multi-tensor factorization (MTF) and present 
the first Bayesian formulation for an extension of joint matrix-tensor factorization.
We also present the first generalized 
formulation of multi-tensor factorization to arbitrary tensors and introduce 
a relaxed low-dimensional decomposition that allows the tensor to factorize 
flexibly. 
Our model decomposes multiple co-occurring matrices and tensors
into a set of underlying factors that can be shared between 
any subset of them, with an intrinsic solution for 
automatic model selection.
Finally, we demonstrate the use of the method in novel coupled matrix-tensor factorization applications, 
including structural toxicogenomics and stimulus-response prediction in neuroimaging.

The rest of the paper is structured as follows: In Section
\ref{sec:prob}, we start by formulating the special case of a single matrix and single tensor 
factorization, inferring components that are shared between both of them, 
or are specific to either one. In Section
\ref{sec:mtf}, we present our Bayesian model that extends to multiple paired
tensors and matrices. In Section \ref{sec:rmtf}
we 
introduce an extension of our new Bayesian solution of Section~\ref{sec:mtf} 
that automatically tunes the decomposition structure for the data. We  
propose a generic formulation in Section \ref{sec:GenericMTF}, and discuss
special cases and related works in Section \ref{sec:relatedworks}.
We validate the performance of our models 
in various technical experiments in Section \ref{sec:exptoy}, and demonstrate
their applicability in a neuroimaging stimulus-response relationship study
and in a novel structural toxicogenomics setting in Section \ref{sec:expreal}. 
We conclude with discussion in Section \ref{sec:discussion}.

%

\textbf{Notations:} \label{sec:notation} We denote a tensor as
$\tensorX$, a matrix as $\X$, a vector as $\textbf{x}$ and a scalar as $x$. 
As presented in \cite{KoldaBader09}, the Mode-1
product $\modeKProdOne$ between a tensor $\tensorA \in
\mathbb{R}^{K\times D\times L}$ and a matrix $\B \in
\mathbb{R}^{N\times K}$ is the projected tensor $(\tensorA \modeKProdOne
\B) \in \mathbb{R}^{N\times D\times L}$, that reshapes the first mode of the tensor. 
A Mode-2 product $\modeKProd$ similarly reshapes the 2nd mode. 
The outer product of two
vectors is denoted by $\outerProd $ and the element-wise product by $*$.
The \textit{order} of a tensor is the total number of axes, modes or ways in the
tensors, while tensor \textit{rank} 
is the smallest number of rank-1 component tensors that generate it; an $N$th order rank one
tensor $\tensorX$ can be presented as $\w_1\outerProd ... \outerProd \w_N$. 
For notational simplicity we present the
models for third order tensors only, including matrices,
for which the dimension of the third mode is one, i.e.
$\tensorX \in \mathbb{R}^{N\times K\times 1}$.


\section{Matrix tensor factorization} \label{sec:prob}

\begin{figure}[t]
     \centering
  \centerline{\includegraphics[width=12cm]{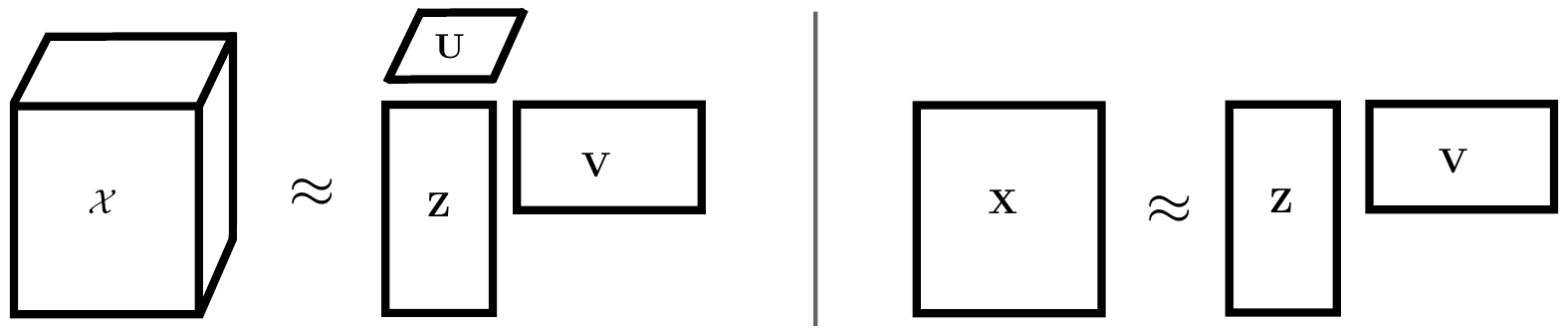}}
     \caption{CP factorization (left) is analogous to matrix factorization (right); each matrix slab of the tensor is just scaled by the corresponding value of each component in an additional matrix $\U$.}
     \label{fig:CPandMF}
\end{figure}

We formulate the joint matrix tensor factorization problem as the identification of a combined 
low-dimensional representation of the matrix 
$\tensorX^{(1)}  \in \mathbb{R}^{N \times D_1 \times 1}$ and the tensor 
$\tensorX^{(2)}  \in \mathbb{R}^{N \times D_2 \times L}$ such that each underlying
factor is either shared by both the matrix and the tensor, or is private to one of them. 
The matrices and tensors can jointly be referred to as different \emph{views} of the 
data, analogously to the terminology used in multi-view learning.
The shared factors represent variation that is common in both the views, while
specific components capture the \emph{view-specific} variation.

The joint factorization can be defined, with
a common set of  low-dimensional latent variables $\Z \in \mathbb{R}^{N \times K}$,
loading matrix $\tensorW^{(1)}  \in \mathbb{R}^{K \times D_1 \times 1}$
and loading tensor $\tensorW^{(2)}  \in \mathbb{R}^{K \times D_2 \times L}$, for each view $t$ as
\begin{align} \label{eqn:matTenFac}
\tensorX^{(t)}  & = \tensorW^{(t)} \modeKProdOne \Z + \eps^{(t)}
\; ,
\end{align}
where $\eps^{(1)} \in \mathbb{R}^{N \times D_1 \times 1}$ is a matrix representing 
noise, while $\eps^{(2)} \in \mathbb{R}^{N \times D_2 \times L}$ is a noise tensor. 
The factorization of the tensor $\tensorX^{(2)}$ into $\tensorW^{(2)} \modeKProdOne \Z$, where $\tensorW^{(2)}$ is unconstrained, 
is equivalent to Tucker-1 factorization \citep{kiers1991hierarchical}, which is analogous to 
matrix factorization of a matricized tensor. 
This factorization still has a huge number of parameters, and the loadings $\tensorW^{(2)}$ can be 
further factorized to model tensorial interactions.
A popular choice is the CP-type formulation which captures trilinear relationships (Fig.~\ref{fig:CPandMF}) and 
has the advantage of being easier to interpret. 
CP is equivalent to a sum of rank-$1$ tensors where each rank-1 tensor is the outer product
of vector loadings in all modes. This rank-$1$ component decomposition of CP 
makes it the choice in most studies. For other useful properties of CP, such as the so-called intrinsic axis property from parallel proportional profiles, see \cite{Cattell44} and \cite{Harshman70}.

Assuming a CP-type decomposition, the loading tensor $\tensorW^{(2)}$ is factorized 
into a tensor product of two latent variable matrices $\V^{(2)} \in \mathbb{R}^{D_2 \times K}$ and $\U^{(2)} \in \mathbb{R}^{L \times K }$. Noting that matrix factorization is a special case of CP,  
both of them can be expressed in the same way, reformulating the joint decomposition as
\begin{align} \label{eqn:matTenFacCP}
\tensorX^{(t)}  & = \sum\limits_{k=1}^K {\z_k \outerProd \vv_k^{(t)} \outerProd \uu_k^{(t)} }  + \eps^{(t)}
\; ,
\end{align}
where the factorization of matrix $\tensorX^{(1)}$ corresponds to Eq.~\ref{eqn:matTenFacCP} with $\U^{(1)}$ as $\mathbf{1}^{1 \times K }$.

A key property of our joint factorization is that each factor can be shared by both the matrix and the tensor, or be specific to either one of them. 
This can be achieved by imposing a group-sparse prior on the loading matrix $\V^{(t)}$ of each view, similar to that in \citep{Virtanen12}. 
The group-sparse prior controls which of the $k$ latent variables are \emph{active} (i.e., non-zero) in each view. 
A component active in both views is said to be shared between them, while a component active in only one captures variation specific to that particular view. 
This formulation allows the matrix and tensor to be decomposed comprehensively, while simultaneously identifying the common and specific patterns.



\section{Multi-tensor factorization } \label{sec:mtf}

We present the first Bayesian treatment of matrix tensor factorization while simultaneously extending it to 
collections of multi-view matrices and tensors, coupled by having a common set of samples. 
The proposed factorization framework is very general, and could as well go by the names collective or collaborative matrix-tensor factorization, or other similar names. Acknowledging that matrices are two-dimensional tensors, we formulate the model family in a simpler but equivalent way as \emph{multi-tensor factorization}.

\begin{figure}[t]
     \centering
     \centerline{\includegraphics[width=12cm]{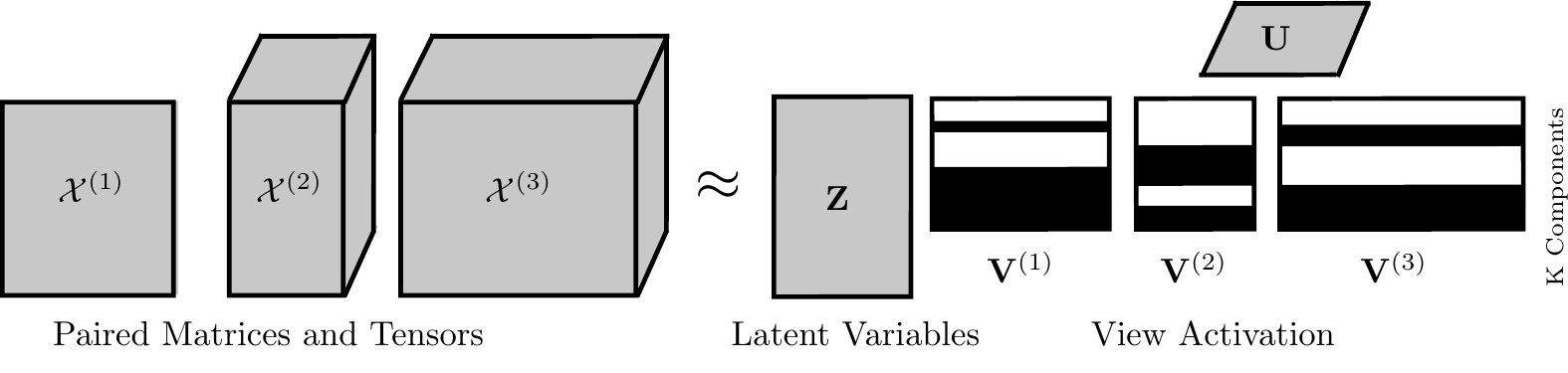}}
     \caption{Multi-tensor factorization projects a set of coupled matrices and tensors into a joint low-dimensional space. In this case, one matrix $\tensorX^{1}$ and two tensors $\tensorX^{2}$ and $\tensorX^{3}$ are decomposed into a common set of latent variables $\Z$ and view-specific projection matrices $\V^{(1,2,3)}$, with the two tensor decompositions having an additional joint set of latent variables $\U$ for the 3rd mode. The $\V$'s control the activity of each component in each view, with black representing a component active in a view, white being switched off. A component active (black) in two or more views captures common patterns of variation.}
     \label{fig:bmmtf}
\end{figure}

Fig. \ref{fig:bmmtf} illustrates the MTF problem for one matrix and two tensors. The samples couple one mode across the collection, and two modes for the tensors. The task now is to perform a joint decomposition of the matrix and the tensors, distinguishing also between the shared and private components. Assuming an underlying CP decomposition for the tensor as in Eq.~\ref{eqn:matTenFacCP}, we perform an unsupervised joint factor analysis and CP-type decomposition of the matrices and tensors, respectively. The joint decomposition is characterized by (i) $\Z$, a common set of latent variables  in all the views (matrices and tensors), (ii) $\U$, the latent variables that model the third mode common to the two \emph{tensor views ($\tensorX^{(2)}$, $\tensorX^{(3)}$) only} and (iii) $\V^{(t)}$, the \emph{view-specific} loadings that control which patterns from $\Z$ and $\U$ are reflected in each view.

This factorization can be seen as a joint FA-CP decomposition where variation patterns can be shared
between matrices and tensors, or be specific to each.
It is motivated by its two main characteristics. 
First, the decomposition of all the matrices and tensors is coupled with the latent 
variables $\Z$ that capture the common response patterns, 
enabling the model to capture dependencies between all the views for 
learning a better factorization. 
Second, the decomposition allows each factor to be active in any combination of the matrices and tensors. 
This gives the formulation the ability 
to capture the underlying dependencies between all or some of the data views, 
as well as to segregate them from the variation that is specific to just one view, often interpretable as (structured) noise.
The dependencies between the views are learned in a fully data-driven fashion, automatically inferring 
the nature of each type of dependency.

Formally, we define multi-tensor factorization for the collection of $t=(1,\ldots,T)$ coupled  tensors and matrices, $\tensorX^{(1)},
\tensorX^{(2)},\ldots, \tensorX^{(T)} \in \mathbb{R}^{N \times D_t
  \times L_t}$, each referred to as a \emph{data view}. 
The central assumption, which will be relaxed later in Section~\ref{sec:GenericMTF}, is that all the views are coupled in one common mode.
Assuming normal distributions and conjugate priors, the generative model underlying the joint matrix-tensor factorization can be expressed as 
\begin{align} \label{eq:mtf}
x^{(t)}_{n,d,l} & \sim \N\left( \z_n^\top(\vv^{(t)}_d \ast \uu^{(t)}_l ) ,\tau_t^{-1} \right) \nonumber\\
z_{n,k}  & \sim \N(0,1) \nonumber\\
v^{(t)}_{d,k} & \sim h_{t,k} \hspace{1mm}  \N\left(0,(\alpha^{(t)}_{d,k})^{-1}\right) + \left(1-h_{t,k}\right)\delta_0  \\
u^{(t)}_{l,k}  & \sim \begin{cases}
    1, & \text{if $L_t=1$},\nonumber\\
    \N(0,1), & \text{otherwise}.
  \end{cases} \\
h_{t,k} & \sim Bernoulli(\pi_k) \nonumber \\
\pi_k & \sim Beta(a^{\pi},b^{\pi}) \nonumber\\
\alpha^{(t)}_{d,k} & \sim Gamma(a^{\alpha},b^{\alpha}) \nonumber\\
\tau_t & \sim Gamma(a^{\tau},b^{\tau}) \nonumber
\; ,
\end{align}
where further constraints can be set depending on the data structure (e.g. $\U^{(2)}=\U^{(3)}$ in the case of Fig.~\ref{fig:bmmtf}). The generative model differs for true tensors ($L_t>1$) and matrices ($L_t=1$) only in terms of the parameter $\U^{(t)}$.
This formulation is similar to
Eq.~\ref{eqn:matTenFacCP}, with the separate noise precision $\tau_t$ for each view. 
For a matrix view, $\tau_t$ samples the noise matrix, and a tensor for a tensor view. 
The Gaussian latent variables $\Z$ are shared across all views, while $\U$ depends on the matrix views only indirectly (via $\Z$). 

The binary variable $h_{t,k}$ controls which components are active in 
each view, by switching the $\vv_{:,k}^{(t)}$ on or off.
This is achieved via the spike and slab prior which samples from a two-part distribution \citep{Mitchell88}.
We center the spike at zero ($\delta_0$), allowing the components to be shut down, 
while the slab is sampled from an element-wise formulation of the ARD prior \citep{Neal96}, parameterized by $\alpha^{(t)}_{d,k}$, 
enabling active components to have feature-level sparsity.
This way the $h_{t,k}$ effectively govern
the sharing of components across all the 
$T$ views, irrespective of whether they are matrices or tensors.
The view-specific loading matrices $\V^{(t)}$
capture active patterns in each data view, while containing zeros for all inactive
components, as illustrated with the white and black patterns in Fig. \ref{fig:bmmtf}. 

The learning of $h_{t,k}$ automatically, in a data-driven way, gives the algorithm 
the power to distinguish between components that are shared between the 
matrices and the tensors, from 
those specific to only one of them. This is achieved in an unbiased fashion by placing an
uninformative beta-Bernoulli prior on $h_{t,k}$ (default parameters being $a^{\pi}=b^{\pi}=1$). 
The formulation also allows the model to  
learn the total cardinality of each dataset, as well as of all the data sets combined. 
This is accomplished by setting $K$ to a large enough value that for a few $k$, $h_{t,k}$
goes to zero in all $t$ views. Such components will be referred to as \emph{empty} components,
and the presence of empty components indicates that $K$ was large enough to model the data.
The effective cardinality of the data set collection is then $K$ minus the number of empty components.
Inference of the model posterior can be done with Gibbs sampling and the implementation is publicly available\footnote{\url{http://research.cs.aalto.fi/pml/software/mtf/}}. The used conjugate priors allow rather straightforward sampling equations, which are omitted here. For the discrete spike and slab prior of $\mathbf{H}$, the sampling was done in a similar fashion as in \cite{Klami13} for Bayesian canonical correlation analysis. The Gibbs sampling scheme for MTF contains inverting a $K\times K$ covariance matrix, resulting in $\mathcal{O}(K^3)$ complexity, which in general makes the model's runtime practical for $K$ in the order of hundreds or less.

For MTF, the uniqueness of the maximum likelihood estimate, and hence asymptotically of the Bayesian solution ($N\rightarrow\infty$), follows from the uniqueness proofs of \cite{Kruskal77,KoldaBader09,sorensen2015coupled}, assuming identifiable structure in the matrices. For full posterior inference with a limited sample size, however, uniqueness is still an open research problem.

\section{Relaxed multi-tensor factorization} \label{sec:rmtf}

\begin{figure}[t]
     \centering
  \centerline{\includegraphics[width=12cm]{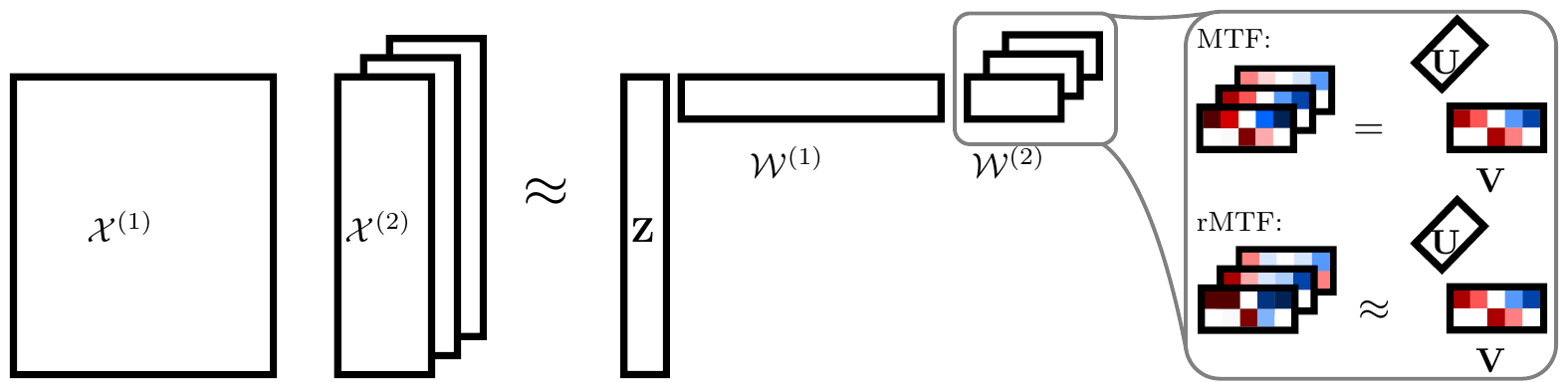}}
     \caption{An illustration of joint factorization of a matrix $\tensorX^{(1)}$ and a tensor $\tensorX^{(2)}$. The generation of the loading matrices corresponding to the tensor, $\tensorW^{(2)}$, is shown in more detail on the right ($D_2=5$ and $K=2$). MTF does a trilinear CP decomposition, as shown by the $\tensorW^{(2)}$ slabs that equal $\V$ with just a scale difference. rMTF allows deviation from this, as illustrated by the slight changes in the $\tensorW^{(2)}$ patterns.
     }
     \label{fig:rbcmtf}
\end{figure}

The trilinear CP structure places a strong assumption on the factorization of a tensor, namely that the projection weights in the second dimension are identical for each slab (3rd mode) with just a varying scale ($\U$, Fig.~\ref{fig:CPandMF}). While in many applications this decomposition has been shown to be useful (see e.g. \cite{acar2007multiway,latchoumane2012multiway}), the assumption may not follow exactly the structure of many data sets. For a more general, relaxed trilinear factorization, we present a novel formulation that allows MTF to capture variation that may not be strictly trilinear.
More specifically, the relaxed formulation aims to decompose the tensor 
$\tensorW^{(t)}$ of Eq.~\ref{eqn:matTenFac} in a flexible
fashion, allowing identification of trilinear structure (characteristic of CP factorization), along 
with structure that is specific to each slab of the tensor (bilinear, matrix structure); but also structures that are along a continuum between the two extremes. This can be seen in Fig.~\ref{fig:rbcmtf}, where the patterns in $\tensorW^{(2)}$ are the same for each tensor slab in MTF, but with varying scale. In contrast, rMTF allows the tensor slabs to deviate from the average patterns.

We formulate a Bayesian solution for the relaxed MTF (rMTF) problem, 
allowing flexible factorization. The
tensors $\tensorX^{t=(1:T)}$
are factorized jointly, capturing factors that are shared
between all, some, or one of the views. 
The formulation is relaxed in the sense that the tensors $\tensorX^{(t)}$ 
have a hierarchical decomposition that allows the model to flexibly
tune between the generic matrix factorization and the trilinear
CP factorization of the tensors. The distributional assumptions of our model are:
\begin{align*} \label{eqn:RMMTF}
x^{(t)}_{n,d,l}   & \sim \N\left( \z_n^\top \w^{(t)}_{l,d} , \tau_{t,l}^{-1}\right) \\ 
w^{(t)}_{l,d,k}   & \sim h_{l,k}^{(t)} \hspace{1mm} \N\left(u^{(t)}_{l,k} v^{(t)}_{d,k}, (\alpha_{d,k}^{(t)})^{-1}  \right) + \left(1-h_{l,k}^{(t)}\right)\delta_0 \\
z_{n,k}, u^{(t)}_{l,k}   & \sim \N(0,1) \\
v^{(t)}_{d,k}      & \sim \begin{cases}
     			   		0, & \text{if $L_t=1$},\nonumber\\
    					\N\left(0,(\beta^{(t)}_{d,k})^{-1}\right), & \text{otherwise}.
  					\end{cases} \\
\alpha^{(t)}_{d,k} & \sim \begin{cases}
     			   		Gamma(a^{\alpha},b^{\alpha}), & \text{if $L_t=1$},\nonumber\\
    					\lambda, & \text{otherwise}.
  					\end{cases} \\
h_{l,k}^{(t)}      & \sim Bernoulli(\pi_k) \\
\pi_k              & \sim Beta(a^{\pi},b^{\pi}) \\
\beta^{(t)}_{d,k}  & \sim Gamma(a^{\beta},b^{\beta}) \\
\tau_{t,l}         & \sim Gamma(a^{\tau},b^{\tau}) \\
\lambda            & \sim Gamma(a^{\lambda},b^{\lambda}) 
\; .
\end{align*}


The key aspect of the model structure is the parameter $\lambda$ describing the similarity of the tensor slabs.
The difference to regular MTF is illustrated in Fig.~\ref{fig:rbcmtf} (right). High $\lambda$ values indicate a factorization that is primarily driven by the low-rank CP structure and is effectively trilinear, whereas low values correspond to highly flexible bilinear multiple matrix factorization. In the tensor formulation, Tucker-1 factorization closely resembles the bilinear multiple matrix factorization, and is considered the least restrictive form of tensor factorization \citep{kiers1991hierarchical}. By default, $\lambda$ will have a relatively uninformative prior ($a^\lambda=b^\lambda=1$), enabling data driven inference of the tensor structure.

Alternatively, informative priors can be considered as well,
if there indeed is some prior information about the structure of the tensors. Additionally, specifying $\lambda_k$ for each component or $\lambda_l$ for each tensor slab would allow learning interesting information about the data structure, namely how strongly each component or slab, respectively, is associated to the trilinear tensor structure. Similarly to MTF, the inference for rMTF is performed with Gibbs sampling, and the implementation is publicly available along with the MTF implementation.




\section{Generalized multi-tensor factorization} \label{sec:GenericMTF}

The notion of coupled matrices and tensors can be formulated for all 
modes of all tensors, allowing decomposition of arbitrarily coupled 
data sets, for investigation of factors shared and specific to each.
To this end, we formulate the general problem of multi-tensor factorization, for data 
collections consisting of matrices and tensors paired in any user-defined fashion. 

\begin{figure}[t]
     \centering
  \centerline{\includegraphics[width=12cm]{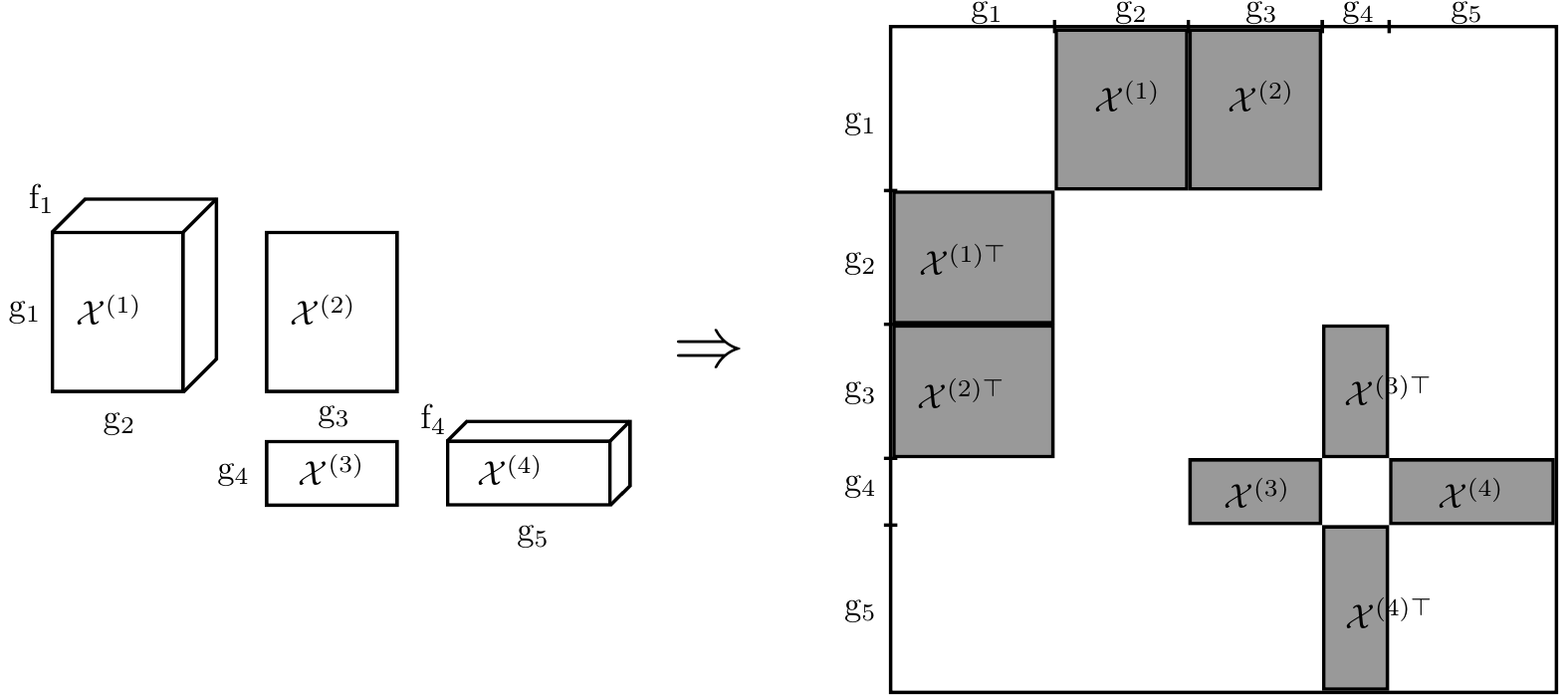}}   
      \caption{Tensor collections with arbitrary pairing between the modes can be assembled into a single large sparse tensor. If the first and the second mode contain groups $\{$g$_1$,...g$_G\}$ and the third mode groups $\{$f$_1$,...f$_F\}$, the whole data collection forms a $\sum_{i=1}^G|$g$_i| \times \sum_{i=1}^G|$g$_i| \times \sum_{j=1}^F|$f$_j|$ tensor. An illustration of the two first modes of this tensor is presented on the right (transpose operating on the first two modes only). A group sparse factorization for this tensor amounts to a MTF generalized to arbitrarily paired tensor collections.}
     \label{fig:GenericMTF}
\end{figure}

The joint factorization task in such 
\emph{multi-mode blocks} can be framed as identification of 
a low-dimensional representation for each of the data 
modes. This is enabled by the observation that the distinction between
samples and dimensions vanishes, as all modes of the data become analogous.
Fig.~\ref{fig:GenericMTF} illustrates the formulation with an example 
of two matrices $\tensorX^{(2)} \in \mathbb{R}^{D_1 \times D_3 \times 1} ,
\tensorX^{(3)} \in \mathbb{R}^{D_4 \times D_3 \times 1}$ and two tensors 
$\tensorX^{(1)} \in \mathbb{R}^{D_1 \times D_2 \times L_1} ,
\tensorX^{(4)} \in \mathbb{R}^{D_4 \times D_5  \times L_4}$,
paired in a non-trivial way. The task in this case is to find 
$K$ factors to represent each of the dimension blocks 
(g$_1$, ..., g$_5$, f$_1$, ..., f$_4$), while capturing the 
common as well as distinct activity patterns that link the data sets 
$\tensorX^{(1)},\tensorX^{(2)},\tensorX^{(3)},\tensorX^{(4)}$ together.
To solve the task, we represent the entire data collection as a tensor 
$\tensorXWH \in \mathbb{R}^{\sum D_{i} \times \sum D_{i} \times \sum L_{i}}$
(partly illustrated in Fig.~\ref{fig:GenericMTF}),
and the $K$ factors as the low-dimensional tensor 
$\tensorW \in \mathbb{R}^{\sum D_{i} \times K \times \sum L_{i} }$,
which has a strict block-structure, active only for regions corresponding 
to data sets being modeled. 

Formally, for $m$ data sets collected into the tensor $\tensorXWH$ the model is
\begin{align*}
\hat{\X}_{:,:,l} & \sim  \W_{:,:,l} \W_{:,:,l}^\top 
\; ,
\end{align*}
where block-structure $\{$g$_1,$g$_2,$g$_3,$g$_4,$g$_5\}$ is imposed by the 
binary variable $h_{b,k,l}$ for $k \in 1 \ldots K$, $l \in 1 \ldots L$
via a spike and slab prior. The relaxed formulation of Section~\ref{sec:rmtf} is
embedded by assuming that the $\sum L_i$ slabs of $\tensorW$ are drawn from the 
mean matrix $\V \in \mathbb{R}^{\sum D_i \times K}$ as

\begin{align*}
w_{d,k,l}      & \sim \begin{cases}
						h_{b_d,k,l} v_{d,k} + (1 - h_{b_d,k,l}) \delta_0, & \text{if $l$ is a matrix-slab}, \\
						h_{b_d,k,l}  \N( v_{d,k} u_{l,k} ,\lambda^{-1}) + (1 - h_{b_d,k,l}) \delta_0, & \text{otherwise}. \\
						\end{cases} \\
v_{d,k}        & \sim \N\left(0,(\alpha_{d,k})^{-1}\right) \\
\alpha_{d,k}   & \sim Gamma(a^\alpha, b^\alpha )
\; ,
\end{align*}
where $b_d$ denotes which group feature $d$ belongs to and the other priors remain unchanged.

The binary variable $h_{b_d,k,l}$ learns in which group each component is active, producing the block-component activations
that extend also to slabs for tensor data sets. The $\lambda$ again controls 
the balance between the trilinear and Tucker-1 structure in the data.
Model specification is completed by assuming normal distribution 
for $\tensorXWH$ and a data view specific noise precision $\tau_t$.

The key characteristic here is that 
group sparsity controls the activation of each latent block-component pair
instead of the data set-component pair; therefore a component's contribution 
in a data set can be switched off in multiple ways. 
For example, in matrix $\tensorX^{(2)}$, the component $k$ can be switched off if either
$h_{1,k,1} = 0$ or $h_{3,k,1} = 0$.
For tensors, the switching notion extends to each of the $L_t$ slabs.
This specification makes the model fully 
flexible and allows components with all possible sharing and 
specificity patterns to be learned, given enough regularization.

This formulation resembles a recent non-negative multiple tensor factorization by \cite{Takeuchi13}. We introduce a Bayesian formulation with relaxed factorization as well as segregate between shared and specific components.

\section{Related work} \label{sec:relatedworks}

The MTF problem and our solution for it are related to several matrix
and tensor factorization techniques. In the following we discuss existing techniques that
solve special cases of the multi-tensor factorization problem, and relate them to our work.

For a tensor coupled with one or more matrices, 
our MTF model can be seen as a Bayesian coupled matrix-tensor factorization (CMTF) method,
which can additionally automatically infer the number and type 
of the components in the data, and enforce feature-level sparsity
for improved regularization and interpretability.
In this line of work, ours is closest to the non-probabilistic CMTF of 
\cite{Acar11,Acar13acmtf,Acar13}. They
assumed an underlying CP decomposition for tensors too,
and used a gradient-based least squares optimization approach.
In their recent work, \cite{Acar13acmtf,acar2014bmc} enforced an $l_1$ penalty 
on the components assuming they can be shared or 
specific to data sets. However, unlike ours, they still required the 
data cardinality ($K$) to be pre-specified. 
Determining the cardinality of
tensors has been considered a challenging problem \citep{KoldaBader09}, 
and our method presents an intrinsic solution for this.
Researchers have also used matrices as side information sources to a 
tensor in CMTF to show improved factorization performance \citep{zheng2012},
while some have also studied underlying factorizations other than the CP, such as the Tucker3 and the block-term decomposition \citep{narita12,sorber2015,Yilmaz11}. 
Recently, solutions have been presented for speeding up the computation
of coupled matrix tensor factorization algorithms on big
data \citep{beutel14,papalexakis14}.
These methods may be generalized to model multi-view matrices and tensors;
however, we present a Bayesian formulation.

When all the tensors have $L_t=1$ and are paired in the first mode, our framework
reduces to the group factor analysis (GFA) problem presented by \cite{Virtanen12}.
GFA has been generalized to allow pairings between arbitrary data modes under the name
collective matrix factorization (CMF) \citep{klami14}, which the formulation in Section~\ref{sec:GenericMTF} generalizes to tensors.

In tensor factorization research, a multi-view problem was recently studied
under the name of multi-view tensor factorization \citep{khan14}. The goal
there was to perform a joint CP decomposition of multiple tensors to
find dependencies between data sets. This method can be seen as
a special case of our model, when all data views are only tensors of the
same order, paired in two modes and assuming a strict CP-type factorization.


\section{Technical demonstration} \label{sec:exptoy}

In this section we demonstrate the proposed MTF methods on artificial data. 
We compare with the multi-view matrix factorization method group factor analysis (GFA)
\citep{Virtanen12}, for which the tensors 
$\tensorX^{(t)} \in \mathbb{R}^{N \times D_{t} \times L_t}$ are transformed into $L_t$ matrices
$\X^{(1)}, \X^{(2)} \ldots \X^{(L_t)} \in \mathbb{R}^{N \times D_{t} }$,
one for each slab of the tensor. In this setting, GFA corresponds to a joint matrix and
Tucker-1 tensor factorization. Thus GFA presents the most flexible tensor
factorization, whereas MTF does a strict CP-decomposition, and rMTF learns
a representation in between these two.

\subsection{Visual example}
\label{sec:vis}

\begin{figure}[t]
\begin{center}
\includegraphics[width=1.0\textwidth]{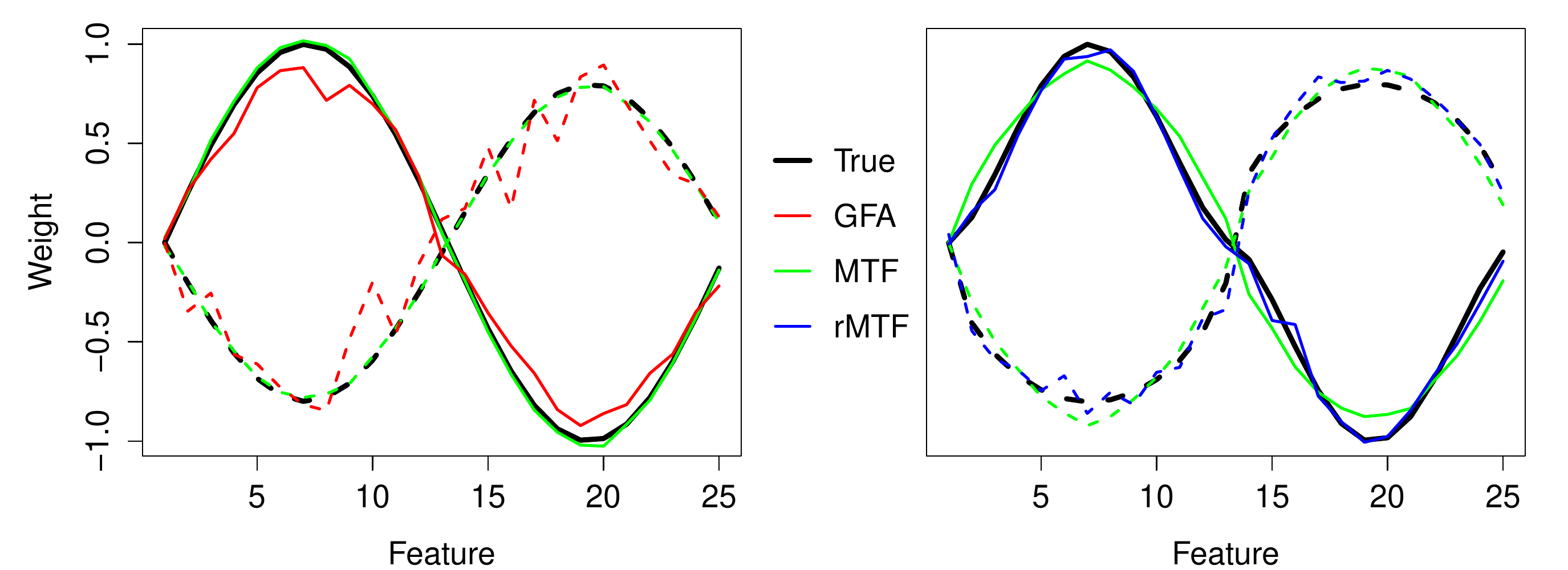}
\caption{Illustration of MTF inference on simulated data when the projections of the tensor are given the shape of a sine wave (``True''). The solid curve denotes the first slab and dashed the second. \textbf{Left:} MTF infers the correct shape more accurately than multi-view matrix factorization (GFA). \textbf{Right:} MTF cannot compensate for tensor slabs deviating from the common signal, whereas the relaxed version (rMTF) can, detecting the true parameters more accurately.}
\label{fig:toyVisual}
\end{center}
\end{figure}

We start with an experiment that illustrates the properties of MTF. We generated $N=300$ data samples from MTF (Eq.~\ref{eq:mtf}), one matrix view with $D_1=50$ and one tensor view with $D_2=50$ and $L=30$. A total of 11 components were used to generate the data: 1 fully shared, 2 specific to the matrix and 8 to the tensor. The fully shared component was given the shape of a sine wave in the second dimension ($\V$), as shown in Fig.~\ref{fig:toyVisual} (left, black curves) for the first 25 features of two tensor slabs. The data set generation was repeated independently 100 times, and the component structures (that is, how many shared and specific components are inferred) detected by the models are shown in Table~\ref{table:comp} (CP). MTF was able to find exactly one shared component in every repetition, but tended to overestimate the number of matrix-specific components (on average 3.37 versus the true 2). GFA was very accurate in detecting the number of matrix-specific components, but overestimated the amount of other structure.
Overall, MTF is able to detect the CP structure considerably more accurately than GFA, which assumes a bilinear structure in the data. The inferred posterior means of a representative data set replication for the shared component weights are shown in Fig.~\ref{fig:toyVisual} (left).
MTF detected the cardinality and component activation correctly, while GFA returned two shared components, out of which the one closer to the true parameters is shown.

To illustrate the advantages of relaxed MTF, we add some distortion to the sine wave of the shared component; the distortion is $v^p$,
where $p=\{0.5, 1.5\}$ for the first two tensor slabs, and an evenly spaced series between 0.3 and 1.7 for the rest. The weights corresponding to powers $0.5$ and $1.5$ are shown in Fig.~\ref{fig:toyVisual} (right, black curves). MTF cannot take into account this kind of differences in the projection weights, whereas rMTF finds a closer match to the correct weights, as shown in Fig.~\ref{fig:toyVisual} (right). In this example, as the trilinear assumption does not hold exactly, MTF returned two shared components, out of which the one closer to the true parameters is shown. It is also worth noting that MTF nevertheless finds a more continuous estimate, as it learns the loadings for all the 30 tensor slabs jointly, instead of allowing individual variance. 
We additionally tested the ability of the models to detect the true component structure with 100 independent repetitions of data generation and model inference. The average number of inferred components is shown in Table~\ref{table:comp} (Relaxed CP) for the three component types: shared, matrix-specific and tensor-specific. As the joint signal is no longer strictly generated from CP, MTF typically uses two shared components to represent it. MTF detects only the number of tensor-specific components, which are strictly CP, more accurately than rMTF. The tested methods overestimate the component numbers somewhat, producing weak spurious components in addition to the stronger true components. Overall, considering the scale of the components as well, they
are fairly accurate in detecting the correct component structure in this simulation study. We also studied how accurately the components are inferred. Table~\ref{table:comp} shows the average (absolute) correlation between a true tensor-specific component and an inferred component that resembles it the most. The Bayesian factorization methods are very accurate in this respect. Due to the CP generation of the tensor-specific components, this holds for MTF in particular.

\begin{table}[t]
\caption{\textbf{Top:} Number of (shared, matrix-specific and tensor-specific) components used to generate the data (``True'') and inferred on average with different models over 100 independent simulated data sets (standard deviation in parentheses). With strict CP data generation, MTF was able to detect the correct number of components shared between the matrix and tensor in every repetition. GFA was more accurate only in detecting the number of matrix-specific components. With the shared component generated from relaxed CP, rMTF was more accurate in detecting the component structure, with the exception of overestimating the number of tensor-specific components. \textbf{Bottom:} The average correlations between the true and the inferred loadings of a tensor-specific component. All the models infer the structure of the specific component very accurately; MTF in particular, as the tensor-specific component has a CP form.}
\begin{tabular}{|c||c|cc|cc|}
\hline 
  & & \multicolumn{2}{c}{CP} \vline & \multicolumn{2}{c}{Relaxed CP} \vline \\\hline 
 & True & MTF & GFA & MTF & rMTF \\\hline 
 Shared & 1 & 1 (0)  & 2.17 (1.57)  & 1.99 (0.1)  & 1.73 (0.98) \\ 
 Matrix & 2 & 3.37 (0.98)  & 2.04 (0.2)  & 3.39 (0.92)  & 1.93 (0.57) \\ 
 Tensor & 8 & 8.19 (0.44)  & 10.75 (1.62)  & 8.07 (0.26)  & 10.07 (1.27) \\ 
\hline Correlation & 1 & 0.9999 (0)  & 0.9954 (0.01)  & 0.9999 (0)  & 0.9937 (0.02) \\ 
 \hline
\end{tabular}
\label{table:comp}
\end{table}

\subsection{Continuum between bilinear and trilinear factorization}

In this experiment, we evaluate the performance of MTF and rMTF on the continuum between multi-view matrix factorization and matrix tensor factorization. The trilinear tensor factorization for slab $l$ of the $t$th tensor is of the form $\sum\nolimits_{k=1}^K u_{l,k}\z_{:,k} \vv_{:,k}^{(t)\top} $, whereas the bilinear multi-view matrix factorization corresponds to $\sum\nolimits_{k=1}^K \z_{:,k} \vv_{:,k}^{(l)\top} $, where the matrix $\V^{(l)}$ is \textit{a priori} independent from all the other data views. We studied the case where neither of these assumptions is correct, but the true factorization is between the assumptions of the two models. For this, we generated $N=15$ samples of training data: one matrix with $D_1=50$ and one tensor with $D_2=50$ and $L=30$. The data set was generated with one fully shared component, 2 specific components for the matrix and 8 for the tensor. The generative model used was a weighted sum of the bilinear and trilinear factorizations. The quality of the models was evaluated by predicting 100 test data samples of one tensor slab ($l=1$) from the rest of the tensor ($l=2,...,30$). For this purpose, we used a two-stage approach: First the parameters were inferred from the fully observed training data, storing the ones that affect the new test samples as well (that is: $\V,\U$ and $\Ta$). In the second stage the latent variables $\Z$ and the missing parts of the test data (in this case tensor slab $l=1$) were sampled given $\{\V,\U,\Ta\}$. This procedure was repeated for all the training phase posterior samples and the final prediction was an average over all the predicted values.

\begin{figure}[t]
\begin{center}
\includegraphics[width=0.5\textwidth]{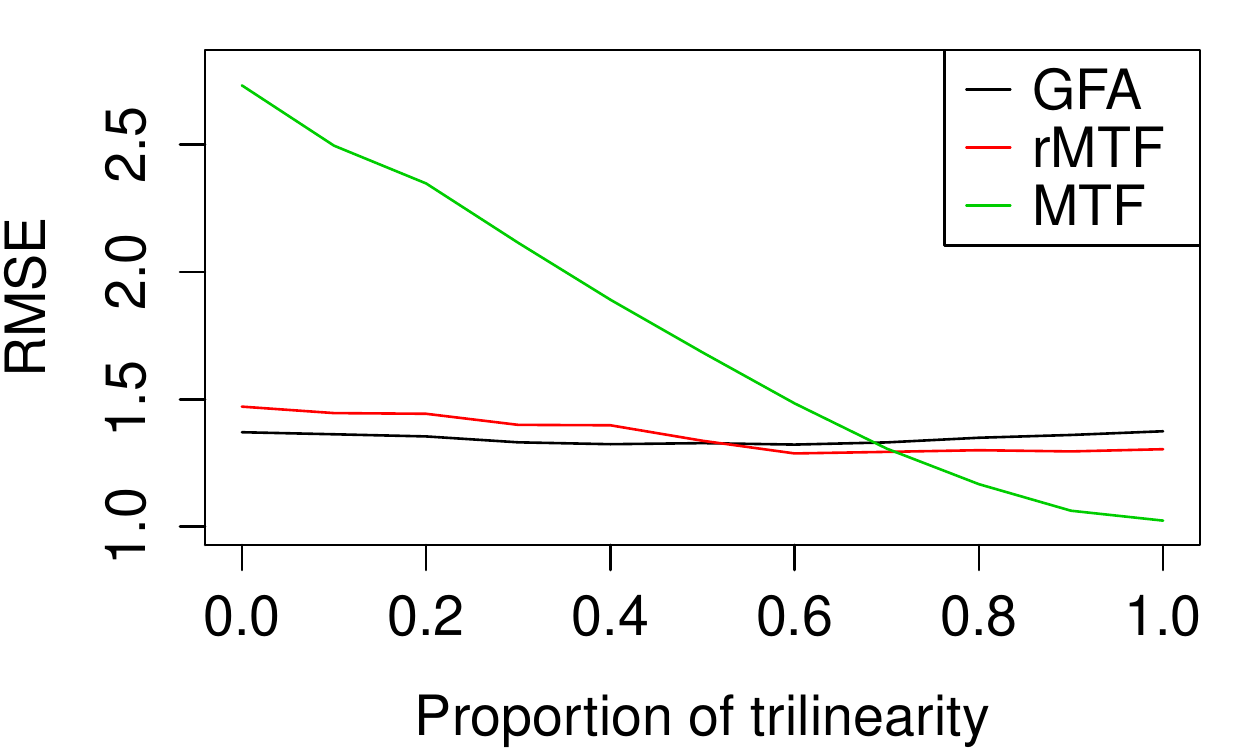}
\caption{A simulation study showing the prediction error (lower is better; an ideal model results in RMSE 1 and a random guess in RMSE 3) as a function of the proportion of trilinearity (vs. bilinearity) in the data; 0 corresponds to (bilinear) multi-view matrix factorization and 1 to perfectly trilinear matrix tensor factorization. As expected, MTF has poor performance when the data do not match the modeling assumptions (on the left), and top-level when they do (on the right). Matrix factorization method GFA is the ideal model when the data have close to bilinear structure, and relaxed MTF is generally close to the better one of these two, making it the most robust model. It suffers somewhat from having a more general parametrization, but is the most accurate model in the mid-region (from 0.55 to 0.7).}
\label{fig:toyTrilinear}
\end{center}
\end{figure} 

The performance of the MTF models and bilinear GFA in this experiment can be seen in Fig.~\ref{fig:toyTrilinear} (averaged over 300 repetitions); the performance was quantified with the prediction RMSE on the left-out test data set. GFA results in a seemingly constant prediction accuracy with respect to the proportion of trilinearity in the data, and is the most accurate model when the data generation process is fully bilinear, as expected. MTF assumes a strictly CP-type of decomposition, and hence varies from weak performance (bilinear data) to ideal performance (trilinear data). Relaxed MTF is the most robust approach, resulting in close to optimal prediction in most of the continuum. When the proportion of trilinearity is in the mid-region, rMTF results in the most accurate predictions. Besides the prediction accuracies, the models' abilities to detect the correct component structure were evaluated as well (data not shown). Even though the number of training samples was very low, the models were able to identify the total number of components rather accurately, but the number of shared components was generally overestimated, underestimating the number of matrix and tensor specific components. In general, prediction accuracy and component detection accuracy were in strong concordance. In the extreme case, with fully trilinear data, MTF was able to detect the exact component structure in 299 out of the 300 repetitions (with 0.9 trilinearity in 290 repetitions). GFA, on the other hand, overestimated the true component amount on average by 0.6, and reported false weak connections between some tensor (or matrix) specific components. The low sample size ($N=15$) suppressed the models' tendencies to report overly many (weak) components, as opposed to the simulation study in Section \ref{sec:vis}.


\section{Applications} \label{sec:expreal}

In this section we demonstrate the use of the proposed MTF methods in two applications: functional neuroimaging and structural toxicogenomics.
To illustrate the strengths of the new methods, we compare them with tensor factorization
methods that are the most closely related to them. 
In particular, we compare with coupled matrix tensor factorization \citep{Acar13}, which decomposes a tensor 
along with a coupled matrix as side information. The available implementation\footnote{\url{http://www.models.life.ku.dk/~acare/CMTF_Toolbox}} uses
CP as the underlying factorization for the tensor, as does our MTF. 
Additionally, we compare against an asymmetric version of coupled matrix tensor factorization (ACMTF)
\citep{Acar13acmtf,acar2014bmc}, which allows
both private and shared components in the data collection. 
CMTF and ACMTF are the closest existing tensor baselines and are non-probabilistic 
formulations.
We also compare our  
method to a multi-view matrix factorization method group factor analysis (GFA)
\citep{Virtanen12} by transforming the tensors 
$\tensorX^{(t)} \in \mathbb{R}^{N \times D_{t} \times L_t}$ into $L_t$ matrices
$\X^{(1)}, \X^{(2)} \ldots \X^{(L_t)} \in \mathbb{R}^{N \times D_{t} }$,
one for each slab of the tensor, as this corresponds to a joint matrix and
Tucker-1 tensor factorization.

Model complexity was determined in a data-driven way, by setting $K$ large enough so that some of the inferred components became shut down. The model parameters $a^{\pi},b^{\pi},a^{\lambda},b^{\lambda}$ were initialized to 1 to represent uninformative symmetric priors. Feature-level sparsity was assumed with parameters $a^{\alpha},b^{\alpha}$ set to $10^{-3}$, while high noise in the data was accounted for by initializing the noise hyperparameters $a^{\tau},b^{\tau}$ for a signal-to-noise ratio (SNR) of 1. All the remaining model parameters were learned. CMTF and ACMTF were run with $K$ values inferred from MTF, as they are unable to learn $K$. ACMTF was run with sparsity setting of $10^{-3}$, as recommended by the authors \citep{Acar13acmtf}. For all the models, the predictions for missing data were averaged over 7 independent sampling chains/runs to obtain robust findings. For missing value predictions, we used the two-stage out-of-sample prediction scheme discussed in the previous section. The data were centered to avoid using components to model the feature means, and unit-normalized to give equivalent importance for each feature.

\subsection{Functional neuroimaging}\label{fmriexp}

A key task in many neuroimaging studies is to find the response related to a stimulus. This is an interesting problem in natural stimulation and multi-subject settings in particular. MTF can be applied in this scenario directly, as the stimulus can generally be represented with a matrix of $N$ samples (time points) and $D_1$ features, whereas the imaging measurements are a tensor with $N$ samples, dimension $D_2$ (e.g. MEG channels) and depth $L$ (subjects). We analyzed a data set presented by \cite{koskinen2014uncovering}, where $L=9$ subjects (one out of ten omitted due to unsuccessful recordings) listened to an auditory book for approximately 60 minutes, while being measured in a magnetoencephalography (MEG) device. In this context, analysis with multi-matrix factorization methods (with subjects regarded as different data views) would assume that the subjects \textit{a priori} do not have any shared information. MTF, on the other hand, aims to decompose the data such that the latent time series (components) have equal feature weights for all the subjects, just scaled differently. Although the imaging device is the same for all the subjects, they will share neither the exactly same brain structure nor functional responses. This makes rMTF a promising model for neuroimaging applications.

The data set was preprocessed in a similar fashion as in \citep{koskinen2014uncovering}. Namely, the 60 minutes of MEG recordings were wavelet-transformed with central frequency 0.54, decreasing the sample size to $N=28547$. The recordings were preprocessed with the signal-space-separation (SSS) method \citep{taulu2004suppression} and furthermore with PCA (jointly for all the subjects) to reduce the dimensionality from 204 (MEG channels) to the number of degrees of freedom left after the SSS procedure ($D_2=70$). As there is a delay in brain responses corresponding to the stimulus, the mel-frequency cepstrum coefficients (MFCC, computed with the Matlab toolbox \textit{voicebox}) describing the power spectrum of the auditory stimulus ($D_1=13$) were shifted to have maximal correlation with the response, and then downsampled and wavelet-transformed to match the MEG recordings.

We inferred the matrix-tensor decompositions with the two methods introduced in this paper and the three comparison methods. The
decomposition was inferred from the $n$ first measurements, and the
models were then used to predict all the later MEG measurements given
the later audio. Relaxed MTF and
GFA were run with $K=500$, leaving empty components with every
training sample size (final component amount ranging from 132 to 380, depending on the sample size). For MTF, even $K=700$ (larger than the total data dimensionality)
was not sufficient, suggesting the data do not fully fit the strong CP
assumptions, and extra components have to be used to explain away some non-CP variation (structured noise). No degeneracy was observed, however, and we used a stronger regularizing prior for MTF
($a^\pi = \frac{1}{b^\pi} = 10^{-3}$ and peaked noise prior for SNR of 1),
ending up with around 500 active components. 
Due to memory requirements, ACMTF was run with only $K=70$, using over 10GB of RAM.
The Bayesian methods inferred with Gibbs sampler were run with 3000 burn-in samples, returning
40 posterior samples with 10 sample thinning for each sampler chain.
We evaluated the convergence of the methods based on the reconstruction of the training data in the posterior samples. Applying the Geweke diagnostic \citep{Geweke92evaluating} under this framework showed that all the chains were converged. The single chain running times of the Bayesian models initialized with $K=500$ were approximately 17 hours with the largest training data (4 hours with 5 minutes of training data). For MTF, initialized with $K=700$, the corresponding running times were roughly 70 and 24 hours.

\begin{figure}[t]
\begin{center}
\begin{tabular}{cc}
\includegraphics[width=0.34\textwidth]{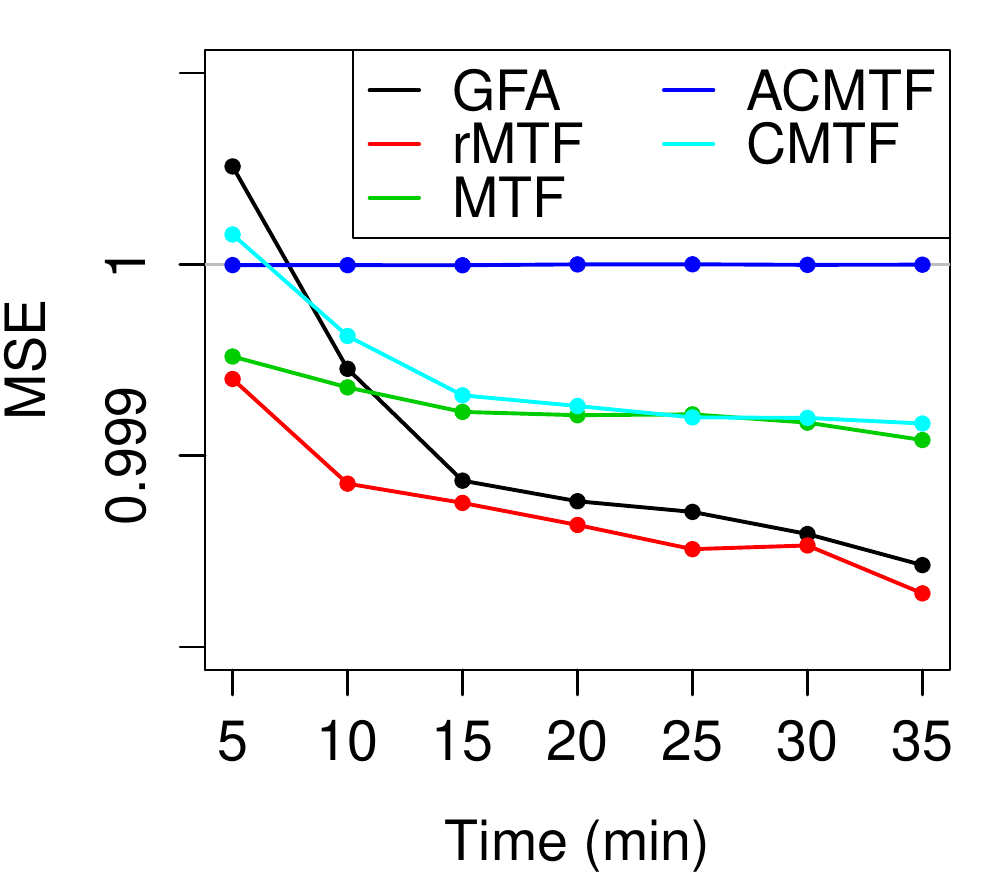} &
\includegraphics[width=0.62\textwidth]{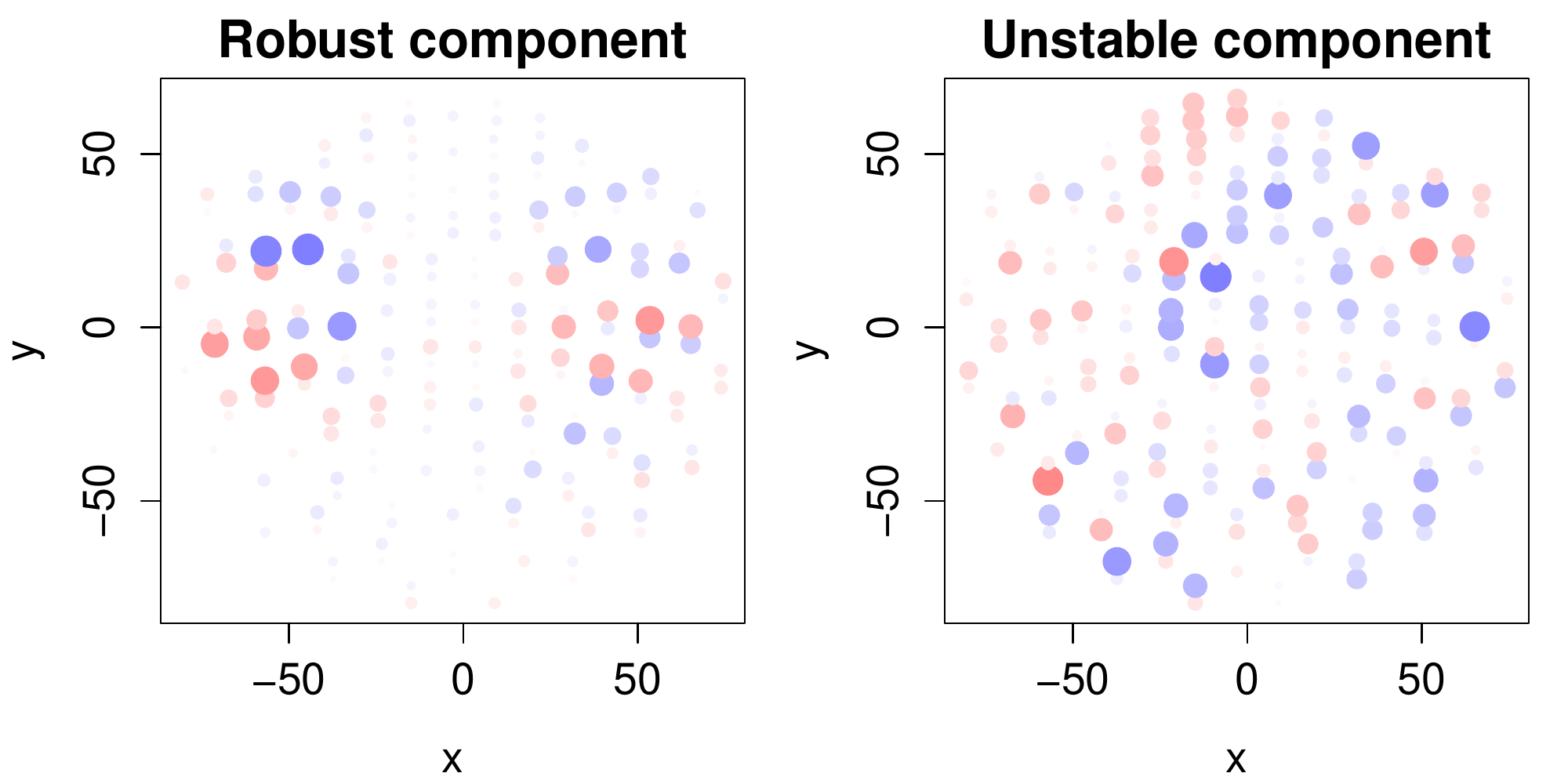}
\end{tabular}
\caption{\textbf{Left:} The mean squared errors (MSE) on predicting (later) brain responses to stimuli as a function of (earlier) training data size (shown as experimental time; ranging from 2500 to 17500 samples). Predicting mean of training data results in MSE of 1. \textbf{Middle:} The feature weights, averaged over subjects, of a robust rMTF component shown in the MEG channel space (top view). Activations are focused around the auditory areas. \textbf{Right:} The averaged feature weights of an unstable GFA component shown in the MEG channel space.}
\label{fig:neuro}
\end{center}
\end{figure}

Relaxed MTF learned the stimulus-response relationship most
accurately for a wide range of training data sizes
(Fig.~\ref{fig:neuro}), whereas the trilinear factorization of MTF
seems to be too strict and the multi-matrix factorization of GFA too
flexible. Despite the challenges in the model complexity determination for MTF, likely due to the overly strict modeling assumptions, it still was able to infer a meaningful factorization. CMTF showed similar performance as MTF once it was given enough
training samples. The sparse solution of ACMTF did not deviate from null prediction
even though different sparsity parameters and convergence criteria were tested.
Relaxed MTF was significantly superior
to all the other methods with all the training set sizes ($p<0.05$; pairwise t-test for each model pair, with MSEs of individual predicted time points as the samples). As the MSEs are close to one and their numerical differences are small, it is worth emphasizing that, especially in natural stimulus settings, the SNR of MEG experiments is very low \citep{koskinen2014uncovering}.
For practical use on data sets with challengingly few samples it is an important
finding that rMTF learned as accurate a stimulus-response relationship
as GFA with roughly half of the training set size on this data set. As overly long
neuroimaging experiments tend to cause decreased signal-to-noise ratio
\citep{hansen2010meg}, relaxed MTF may offer significant benefit in this area. 

The most robust finding of the factorization models in this application is the brain response to the energy of the speech signal. Of the 13 acoustic MFCC features, 9 are highly dependent on the signal energy and hence two-peaked, corresponding to words and breaks between the words. A robust component found in all the rMTF chains had similar two-peaked structure, and was found to be active in the auditory areas of the brain (Fig.~\ref{fig:neuro}, middle). With enough samples, GFA was able to detect this component robustly as well, but it produced more unstable components present in individual sampling chains only. These components had no clear structure in the MEG channels, as shown in Fig.~\ref{fig:neuro}, and are hence likely to be artifacts explaining noise in the recordings. No other robust shared components were found with rMTF, likely because we analyzed the relationship between the stimulus and the brain response at only one time lag. Various brain responses occur at different lags; in this experiment we focused on the initial response, simple auditory processing of the heard sound. For a more thorough neuroscientific analysis it would be important to take the temporal nature of the events more directly into account. Besides the typically 1 to 4 fully shared components (one robust over most of the chains), rMTF and GFA typically had 1 to 4 components specific to the acoustic features and around 200 describing the MEG measurements, either active for all the subjects, or for a subset of them. The wide range of MEG-specific components describe brain activity unrelated to the task, or not sufficiently described by the acoustic features. From the experimental perspective of finding stimulus-related activity, these components can be thought to describe structured noise.

\subsection{Structural toxicogenomics}

\begin{figure}[t]
     \centering
  \centerline{\includegraphics[width=10cm]{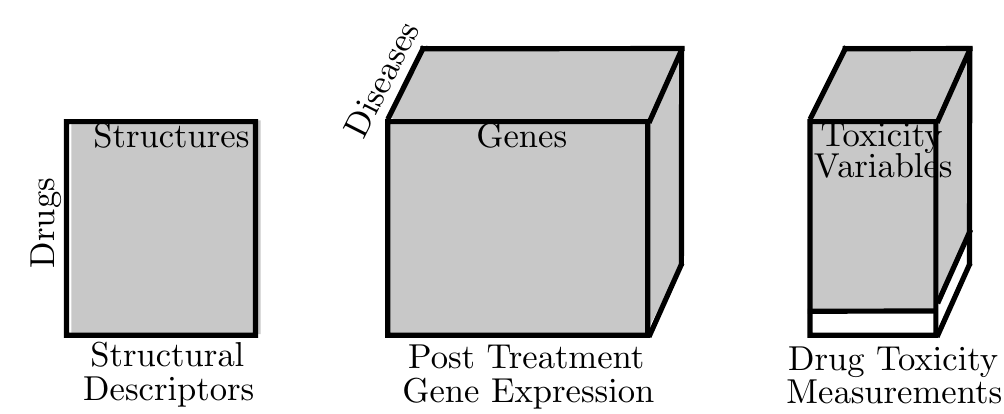}}   
\vspace{3mm}
     \caption{Structural toxicogenomics data set with a matrix of drug descriptors paired with a tensor of gene expression responses of different diseases to the drugs, and the corresponding toxicity profiles.}
     \label{fig:ToxGenData}
\end{figure}

We next analyzed a novel drug toxicity response problem, where the tensors
arise naturally when gene expression responses of multiple drugs are
measured for multiple diseases (different cancers) across the
genes. The data contain three views, the structural descriptors of drugs (a matrix),
measurement of post-treatment gene expression (a tensor), 
and drug toxicity (a tensor) as shown in Fig.~\ref{fig:ToxGenData}.
As drugs have several diverse effects on cells, they can be hypothesized to have been generated
by underlying factors, some of which may be common across all diseases while others specific to only few. 

In this setting, MTF can be used
to answer two key questions: (1) which parts of the responses are
specific to individual types of cancer and which occur across cancers,
and which of these responses are related to known structural properties of the drugs; 
and (2) can we use the links from gene expression responses along with structural properties
of drugs, to predict toxicity of an unseen drug.

For the first problem, the response patterns of drugs, if uncovered, 
can help understand the mechanisms of toxicity
\citep{Hartung12}. The identification of links from structural properties 
of drugs (a matrix) to the gene/toxic responses (tensors) opens up the 
opportunity for drug-designers to better understand the functional 
effects of drugs' structural fragments \citep{KhanBio14}. Secondly, 
it is interesting to explore how well our method predicts unseen responses
by using multiple side-information sources and the structure in the data.

\begin{figure}[t]
\begin{center}
\centerline{\includegraphics[width=9cm]{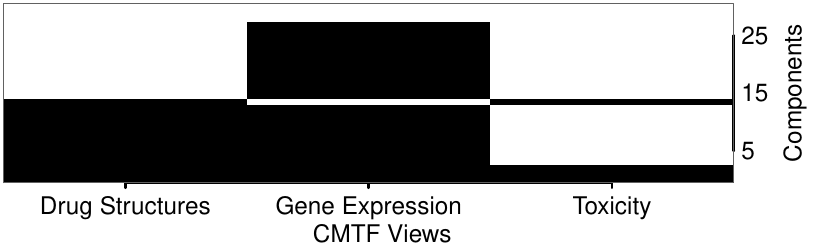}}
\caption{Component activity plot for the structural-toxicogenomics data set (black: active, white: all zeros). 
 The model found 3 interesting components active in all the views, capturing patterns 
 shared by structures  of drugs, their disease-specific gene expression responses, and the toxicity 
 measurements. These components capture the toxic responses of drugs.
}
\label{Fig:toxGenACT}
\end{center}
\end{figure} 

The data set contained three views (Fig.~\ref{fig:ToxGenData}). The first contained
structural descriptors of $N=73$ drugs. The descriptors known as functional
connectivity fingerprints FCFP4 represent the presence or absence of a 
structural fragment in each compound. For this data set the drugs are described
by $D_1=290$ small fragments, forming a matrix of $73$ drugs by $290$ fragments. 
The second view contains the
post-treatment differential gene expression responses $D_2=1106$ of
$N=73$ drugs, as measured over multiple diseases or here cancer types, $L=3$. The
third view contained the corresponding drug sensitivity
measurements, $D_3=3$. The two tensors are paired
with the common identity of $N=73$ drugs and $L=3$ cancer types, while 
the drug structure matrix is paired with the tensors on the common set of $N=73$ drugs.
The gene expression data were obtained from the
ConnectivityMap \citep{Lamb06} that contained response measurements of
three different cancers: Blood Cancer, Breast Cancer and Prostate
Cancer. The data were processed so that gene expression values
represent up (positive) or down (negative) regulation from the
untreated (base) level. Strongly regulated genes were selected,
resulting in $D_2=1106$. The Structural descriptors (FCFP4) of the drugs
were computed using the Pipeline Pilot software\footnote{\url{http://accelrys.com/products/pipeline-pilot/}} 
by Accelrys. The drug screen data for the three cancer
types were obtained from the NCI-60 database \citep{Shoemaker06},
measuring toxic effects of drug treatments via three different
criteria: GI50 (50\% growth inhibition), LC50 (50\% lethal
concentration) and TGI (total growth inhibition). The data were
processed to represent the drug concentration used in the connectivity
map to be positive (when toxic) and negative indicating non-toxic. 
The methods were run with 5000 burn-in samples, 
returning 40 posterior samples with 10 sample thinning for each sampler chain. 
Convergence was examined as in Section~\ref{fmriexp}, and all except $\sim$25\% of the rMTF chains had converged.
The single chain running times of the Bayesian models initialized with $K=30$ were around 1 hour.

MTF resulted in 3 components shared between the 
structural descriptors, the gene expression and toxicity views, 
revealing that some patterns are indeed shared (Fig.~\ref{Fig:toxGenACT}). 
Model complexity was again selected, as in the previous section, by assigning  $K$ large enough, here $K=30$, such that then the sparsity prior shuts some of them off (to zeros). 
The 3 shared components can be used to form hypotheses about 
underlying biological processes that characterize toxic responses of drugs, and we 
find all three of them to be well linked to either established biology or potentially 
novel findings.

\begin{figure}[t]
\begin{center}
\begin{tabular}{cc}
\includegraphics[width=0.3\textwidth]{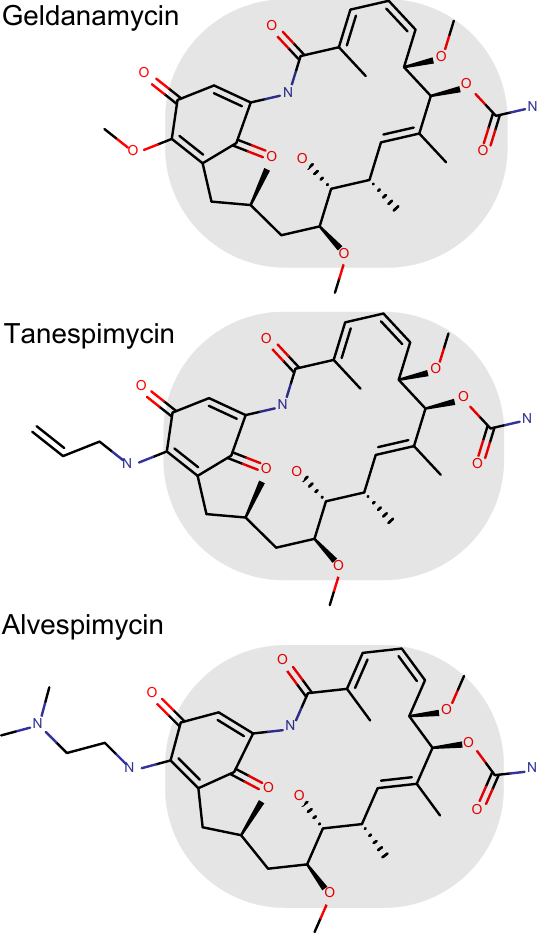} &
\includegraphics[width=0.6\linewidth]{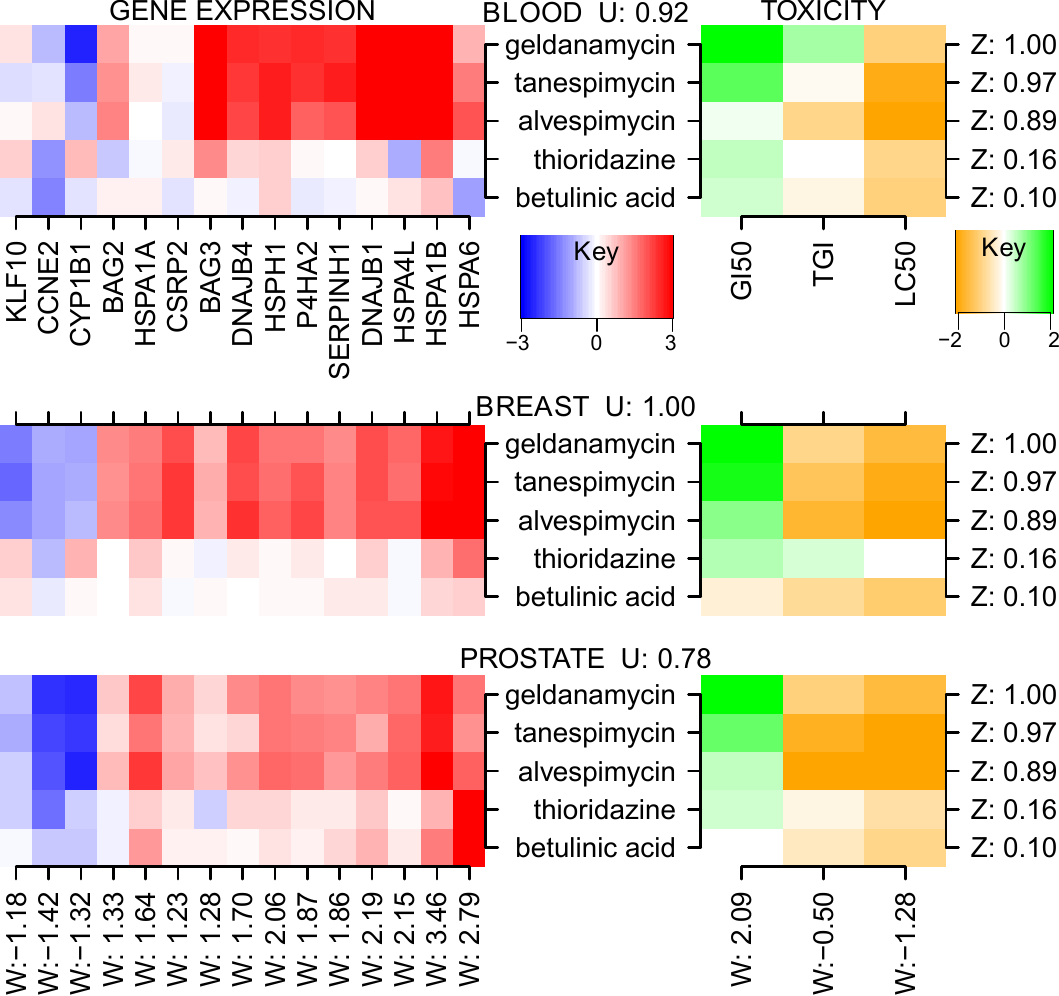} %
\end{tabular}
\caption{An illustrative example: Component 1 captures the well-known heatshock protein
  response. The model identifies the common structural descriptors of the drugs (left)
 as drivers of the biological (middle) and toxic (right) responses. The responses
 are shown for the top genes and all toxicity variables, and top drugs for all the three cell lines (rows), along with the $\Z, \U$ and $\W$ loadings (shown at the ticks). The top genes and drugs are identified as those with the highest loadings in $\W$ and $\Z$ respectively.
 This component links structures of the heatshock protein inhibitor drugs (circled with gray on the left)
with strong upregulation of the heatshock protein genes (red) to high toxicity (green).}
\label{Fig:toxGenVIS}
\end{center}
\end{figure}

The first component captures the well-known heatshock protein
response, of the three HSP90 inhibitor drugs (Fig.~\ref{Fig:toxGenVIS}, left). 
The response is characterized by a strong upregulation of
heatshock genes in all three cancers (Fig.~\ref{Fig:toxGenVIS}, middle) and
the corresponding high toxicity indications in GI50
(Fig.~\ref{Fig:toxGenVIS}, right). The component 
identifies similarities between the three close structurally analogous drugs, 
which is in line with knowledge that the drugs directly bind to the HSP90 protein \citep{stebbins1997crystal}.
The heatshock protein inhibition response has already been well studied 
for treatment of cancers \citep{Kamal03}, evaluating its potential 
therapeutic efficacy. This trilinear MTF component could
have been important in revealing the response, had the mechanism not already been discovered.

The second component captures DNA damage response of several structurally similar 
cardiac glycoside drugs and a structurally different drug, bisacodyl, which is a laxative. 
Interestingly, our component found the response of bisacodyl to be specific to 
only one of the cancer types. The link of bisacodyl with cardiac glycosides
has very recently been found \citep{iorio2010}, but the possible cancer specificity, which comes out naturally with our approach, is new.

The third and final shared component captures a common response of protein synthesis 
inhibitors along with an anti-metabolite (8-azaguanine) drug. Interestingly, the response 
is specific to two of the three cancer types, namely blood and prostate cancers. With 
8-azaguanine having been used in blood cancer before \citep{colsky1955}, our component
opens up an interesting opportunity for its exploration in prostate cancer.

For completeness, we also examined the first view-specific component, which captured a 
non-toxic response of several sequence-specific DNA binding transcription factor genes. 
The response is driven by two drugs, nystatin and primaquine. Nystatin is an anti-fungal drug 
while primaquine (an anti-malarial drug) is already well-established for use in the treatment 
of fungal pneumonia \citep{Noskin92}.

This application demonstrated the model's ability to identify both established and novel
links between multiple views. Systematic studies along these lines could be very 
valuable in precision medicine 
for targeting specific disease types, and in drug design when tailoring drugs to 
match a desired response profile.

\begin{table}[!t]
\setlength{\tabcolsep}{8pt}
\caption{Structural toxicogenomics: toxicity prediction of an unseen drug given its structural descriptors and genomic responses. Average Prediction RMSE is given over the entire set of drugs, with lower values signifying better performance. A paired t-test over the prediction RMSE of these drugs was carried out to test the significance of the difference between the reported methods. MTF outperforms GFA, CMTF and ACMTF significantly with t-test p-values $< 0.02$.}
\vspace{3mm}
\label{tab:toxgen}
\begin{center}
\begin{tabular}{|l|cc|ccc|}
\hline
& \textbf{MTF} & rMTF & CMTF & ACMTF & GFA \\
\hline
Mean & \textbf{0.579} & 0.584 & 0.692 & 0.727  & 0.642  \\
StdError & \textbf{0.062} & 0.064 & 0.079 & 0.088 &  0.070 \\
\hline
\end{tabular}
\end{center}
\end{table}

We next evaluated the model's ability to predict the toxicity response of a new drug, by 
using the gene expression data (tensor) and the structural descriptors (matrix). 
Both are used as side information sources, coupled in a multi-view setting. 
This is done by modeling the 
dependencies between all the observed data sets and then using the learned dependencies 
to predict the toxicity response of a new drug, given the side information sources. The entire toxicity slab 
(shaded white in Fig.~\ref{fig:ToxGenData}) for each drug is predicted
using its gene expression and structural descriptors. 
We compared with the existing methods GFA, CMTF and ACMTF as baselines. 
GFA was run by transforming the gene expression and toxicity tensors into matrices, one for each of the $L=3$ slabs.
We performed leave-one-out prediction and report the
average prediction error of unseen drugs (RMSE) in Table \ref{tab:toxgen}.
The results demonstrate that MTF predicts drug toxicity of unseen 
drugs significantly better, confirming that it solves well the task for which it was designed. 

\section{Discussion} \label{sec:discussion}

We introduced Bayesian multi-tensor factorization (MTF) and
as its special case the first Bayesian 
formulation of joint matrix-tensor factorization, extending the former formulation further 
to multiple sets of paired matrices and tensors. Our model
decomposes the data views into factors that are shared
between all or some of the views, and those 
that are specific to each. It also learns the total number 
and type of the factors automatically for each data collection.

We simultaneously extended our novel formulation to explore a relaxed
underlying tensor factorization problem,
automatically moving between the CP-type of a trilinear model
and a generalized variant of a Tucker-1 type of decomposition.
The CP and the Tucker-1 decompositions 
fall out as special cases of the relaxed variant.
This is important as Tucker-1, in particular, is suitable when data sets 
have minimal trilinear structure, while 
the CP-type trilinear decomposition in this paper has the advantage of being 
interpretable analogously to the matrix factorizations more familiar 
to most analysts.

We validated the models' performances in identifying the correct 
components on simulated data, and illustrated that the relaxed
factorization performs well when the structure of the data is unknown
or not strictly CP or Tucker-1 type. 
The models' performances were then demonstrated on a new
structural toxicogenomics problem and on stimulus-response relationship analysis 
in a neuroimaging experiment, yielding
interpretable findings matching with some expected effects and recent discoveries, and including also potential new 
innovations. The experiments indicated that taking the appropriate structure of the 
data into account makes the results both more accurate and easily interpretable.

Our work opens up the opportunity for novel applications and 
integrative studies of diverse and partially coupled data views, 
both for predictive purposes and feature identification. 



\subsection*{Acknowledgments} 
This work was financially supported by the Academy of Finland 
(Finnish Centre of Excellence in Computational Inference Research 
COIN, grant no 251170; and ChemBio project, grant no 140057)
and the Finnish Graduate School in Computational Sciences (FICS).
We would like to thank Miika Koskinen for
his help with the neuroscience application, and Juuso Parkkinen for providing the toxicity profiles. 
The calculations presented above were performed using computer resources within the Aalto University School of Science ``Science-IT" project.

\noindent Conflict of Interest: The authors declare that they have no conflicts of interest.

\bibliographystyle{spbasic}
\bibliography{manuscript}

\end{document}